\documentclass[11pt]{article}
\usepackage[T1]{fontenc}\usepackage{graphicx,array,booktabs,caption,xcolor,amsmath,amssymb,url}
\usepackage[hidelinks]{hyperref}
\newsavebox{\ResultTableBox}
\title{SoccerSynth-Detection: A Synthetic Dataset for Soccer Player Detection}
\author{Haobin Qin$^{1}$, Calvin Yeung$^{1}$, Rikuhei Umemoto$^{1}$, and Keisuke Fujii$^{1,2,3}$\\[4pt]
\small $^{1}$Graduate School of Informatics, Nagoya University, Nagoya, Aichi, Japan\\
\small $^{2}$RIKEN Center for Advanced Intelligence Project, Suita, Osaka, Japan\\
\small $^{3}$PRESTO, Japan Science and Technology Agency, Kawaguchi, Saitama, Japan\\
\small Corresponding author: \texttt{fujii@i.nagoya-u.ac.jp}}
\date{}
\begin{document}
\maketitle
\begin{abstract}
Annotating players in broadcast soccer footage is costly. We present SoccerSynth-Detection, a dataset of 54,673 simulator-generated images with player bounding boxes, and evaluate it with YOLOv8n, YOLOv8l, and RT-DETR-L on SoccerNet-Tracking and SportsMOT soccer. For target-domain training, we hold out entire matches or source videos for validation and compare training procedures across three matched seeds. Adding an equal number of synthetic images to the real training set improves mean AP50:95 by 1.43--3.38 percentage points over real-only training across the six detector--dataset combinations. Synthetic pretraining followed by real-data fine-tuning improves mean AP50:95 by 0.22--3.37 points. Mixed training also outperforms the three tested real-image augmentation controls on this metric in every combination. Direct transfer varies by detector and target, and the best tested proportion of synthetic data depends on both.
\end{abstract}
\section{Introduction}\label{Introduction}
Soccer-player detection serves as a foundational task in soccer video processing, directly influencing subsequent tasks such as tracking \cite{hu2024basketball,scott2022soccertrack}, action recognition \cite{vanderplaetse2020improved,giancola2018soccernet,giancola2021temporally,wu2022survey,yin2024survey}, action prediction \cite{fang2024foul}, and game state reconstruction \cite{somers2024soccernet,ren2010multi}. Its accuracy is shaped not only by detector design but also by the scale, diversity, and accessibility of annotated training datasets.

Traditional detection methods such as the Viola-Jones Detector \cite{viola2001rapid}, Histograms of Oriented Gradients \cite{dalal2005histograms}, and Discriminatively Trained Part-Based Models \cite{felzenszwalb2009object} rely on manually designed features. In contrast, deep learning-based object detectors have developed along several lines, including two-stage detectors such as Faster R-CNN \cite{girshick2015fast}, one-stage real-time detectors such as the YOLO family \cite{redmon2016you,redmon2017yolo9000,redmon2018yolov3,bochkovskiy2020yolov4,jocher2021ultralytics,li2022yolov6,wang2023yolov7,varghese2024yolov8}, and real-time transformer-based detectors such as RT-DETR \cite{zhao2023detrs}.

For soccer-player detection, the scarcity of annotated broadcast-style soccer video data remains a critical bottleneck for applying modern detectors. Compared with broader visual recognition tasks, soccer-player detection datasets are difficult to construct and redistribute at scale because match videos are often subject to copyright restrictions, and player annotations are costly under frequent occlusions, dynamic backgrounds, lighting variations, and motion-induced blurring. Among various sports, soccer stands out with its large number of players, which exacerbates occlusions and complicates tracking. There are several datasets for the detection of team sports players, such as SoccerNet-Tracking \cite{cioppa2022soccernet}, a tracking dataset from live soccer match recordings; SoccerTrack \cite{scott2022soccertrack} and TeamTrack \cite{scott2024teamtrack}, using fisheye cameras and drones to cover various sports; SoccerDB \cite{jiang2020soccerdb}, a target detection dataset from TV broadcasts; 3x3 basketball datasets \cite{yin2024enhanced,hu2024basketball}; and SportsMoT \cite{cui2023sportsmot}, designed for multi-sport player tracking.

Employing simulators to generate synthetic datasets has emerged as a promising solution to address dataset scarcity, gaining significant attention in recent years for enabling large-scale, low-cost, and precise label generation across domains such as pedestrian detection \cite{fabbri2021motsynth,hattori2015learning,kohl2020mta}, autonomous driving \cite{sun2022shift}, and indoor scene understanding \cite{roberts2021hypersim,zheng2020structured3d}. Recent VR and digital-twin studies also show that controllable virtual environments can support synthetic data generation and person-related visual analysis \cite{mukhopadhyay2021generating,mukhopadhyay2022virtual,mukhopadhyay2021validating}. In the field of soccer, several studies have attempted to create synthetic datasets. For instance, the SoccerNet-Depth dataset \cite{leduc2024soccernet} employs NVIDIA Nsight Graphics to extract depth maps of players from eFootball games. The RADepthNet study \cite{li2022radepthnet} produced a soccer depth map dataset known as the FIFADataset, using the open-source tool RenderDoc. Web-based tools are utilized to generate soccer field datasets for keypoint extraction \cite{fernandes2022identifying}. Furthermore, Kaneko et al. \cite{kaneko2024augmenting} implemented the Google Research Football simulator \cite{kurach2020google} to train algorithms aimed at predicting pass movements in soccer. Additionally, Qin et al. \cite{qin2022end} outline a methodology that employs a soccer camera simulator with reinforcement learning models. However, despite these efforts, publicly released synthetic datasets providing bounding-box annotations specifically for soccer-player detection from a single main broadcast-camera viewpoint remain limited.

To address this gap, we build on the soccer simulator developed in our previous work \cite{qin2022end} and introduce \textit{SoccerSynth-Detection}, a controllable synthetic dataset and generation pipeline specifically designed for soccer-player detection. We target player detection in soccer match videos captured from a single main broadcast-camera viewpoint, represented by the center-camera configuration of our simulator. SoccerSynth-Detection provides scalable and automatically annotated player samples with controllable variations in lighting, textures, player configurations, and visual conditions. We evaluate SoccerSynth-Detection using YOLOv8n, YOLOv8l \cite{varghese2024yolov8}, and RT-DETR-L \cite{zhao2023detrs} on SoccerNet-Tracking and the soccer sequences of SportsMoT because their broadcast views are aligned with this target setting \cite{cioppa2022soccernet,cui2023sportsmot}. SoccerTrack and TeamTrack were excluded from quantitative evaluation because their fisheye and drone full-pitch views differ substantially from the viewpoint modeled by SoccerSynth-Detection \cite{scott2022soccertrack,scott2024teamtrack}.

We compare direct transfer, synthetic pretraining followed by real-data fine-tuning, and joint training on real and synthetic images. We also test real-image augmentation controls and vary the proportion of synthetic training data.

\section{Method}
\label{sec:3.1}

\subsection{Simulator Construction}
\label{sec:3.1.1}

\begin{figure}[!htbp]
\centering
\includegraphics[width=.55\linewidth,height=.28\textheight,keepaspectratio]{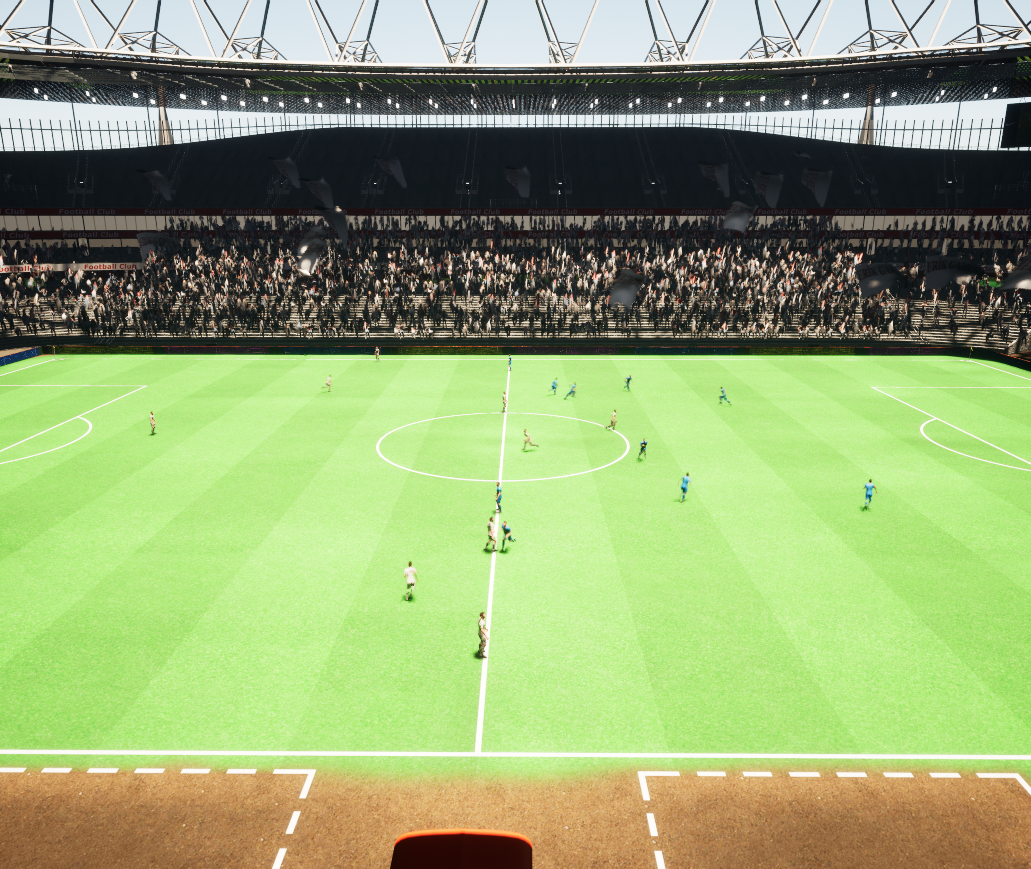}
\caption[Main broadcast-camera view in SoccerSynth.]{Center-camera view used for broadcast-style player-detection images.}
\label{fig:3.1}
\end{figure}

To generate player-detection images from the main broadcast viewpoint, we extended the existing soccer stadium simulator \cite{qin2022end} with a center-camera capture system. The camera positions, viewing angles, and field-of-view settings were configured with reference to SoccerNet-Tracking \cite{cioppa2022soccernet} and the soccer subset of SportsMOT \cite{cui2023sportsmot}, as illustrated in Figure~\ref{fig:3.1}. We also incorporated dynamic spectator models purchased from the Unreal Engine Marketplace to enrich the stadium background.

To generate varied player poses and spatial interactions, we integrated character models and animations sourced from the Unreal Engine Marketplace and implemented player control using Behavior Tree and artificial intelligence (AI) Controller components in Unreal Engine. The simulator runs autonomous match-like sequences in a 4-4-2 formation. Players pass, dribble, shoot, receive, and reposition according to role-based rules, while the defending player nearest to the ball attempts to regain possession. The physics engine simulates ball movement during passing and receiving. The scene is reset when the ball exits the field or the simulated match time expires.

\subsection{Data Generation}
\label{sec:3.1.2}

We applied domain randomization during rendering to broaden visual variation, with the aim of improving detector robustness on real images. Following prior work on synthetic dataset generation \cite{fabbri2021motsynth,hattori2015learning,kohl2020mta,sun2022shift,roberts2021hypersim,zheng2020structured3d}, we randomized lighting, field materials, player appearance, camera orientation, and camera field of view. Frames and bounding boxes were captured using the ``2D Bounding Box for Deep Learning Image Detection'' plugin purchased from the Unreal Engine Marketplace, with documentation provided by the Unreal 2D Bounding Box Plugin project \cite{plasma}. The annotation targets are the 22 on-field players, although the number of exported bounding boxes varies by frame. The plugin application programming interface (API) returns full-body bounding boxes: occlusion does not reduce the annotated extent to the visible part of the body. Annotations are stored in You Only Look Once (YOLO) format.

\begin{figure}[!htbp]
\centering
\begin{minipage}[b]{0.32\linewidth}\centering
\includegraphics[width=\linewidth,height=0.25\textheight,keepaspectratio]{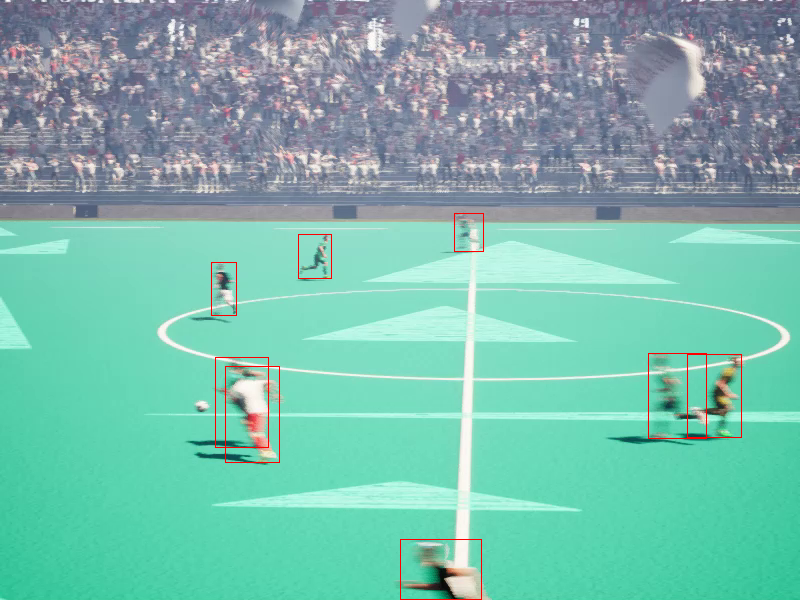}\par\smallskip{\small (a) Motion blur}
\end{minipage}
\hfill
\begin{minipage}[b]{0.32\linewidth}\centering
\includegraphics[width=\linewidth,height=0.25\textheight,keepaspectratio]{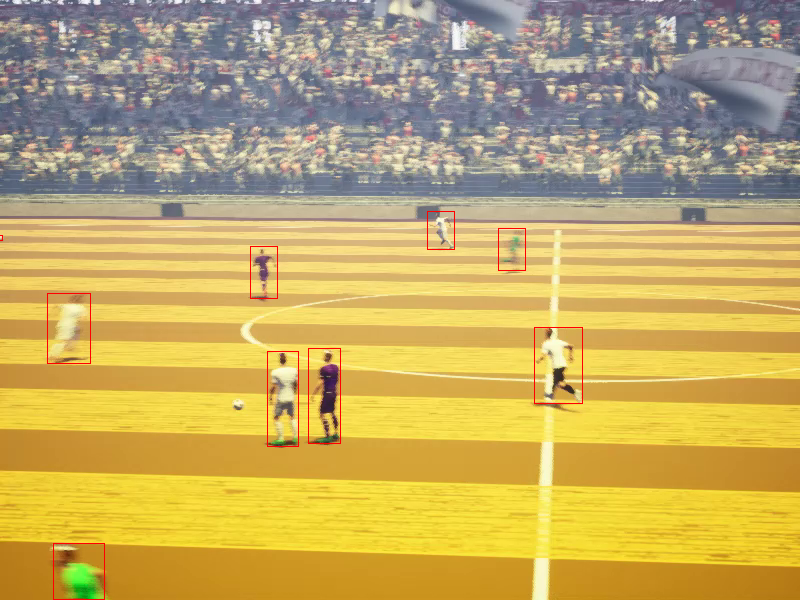}\par\smallskip{\small (b) Field appearance}
\end{minipage}
\hfill
\begin{minipage}[b]{0.32\linewidth}\centering
\includegraphics[width=\linewidth,height=0.25\textheight,keepaspectratio]{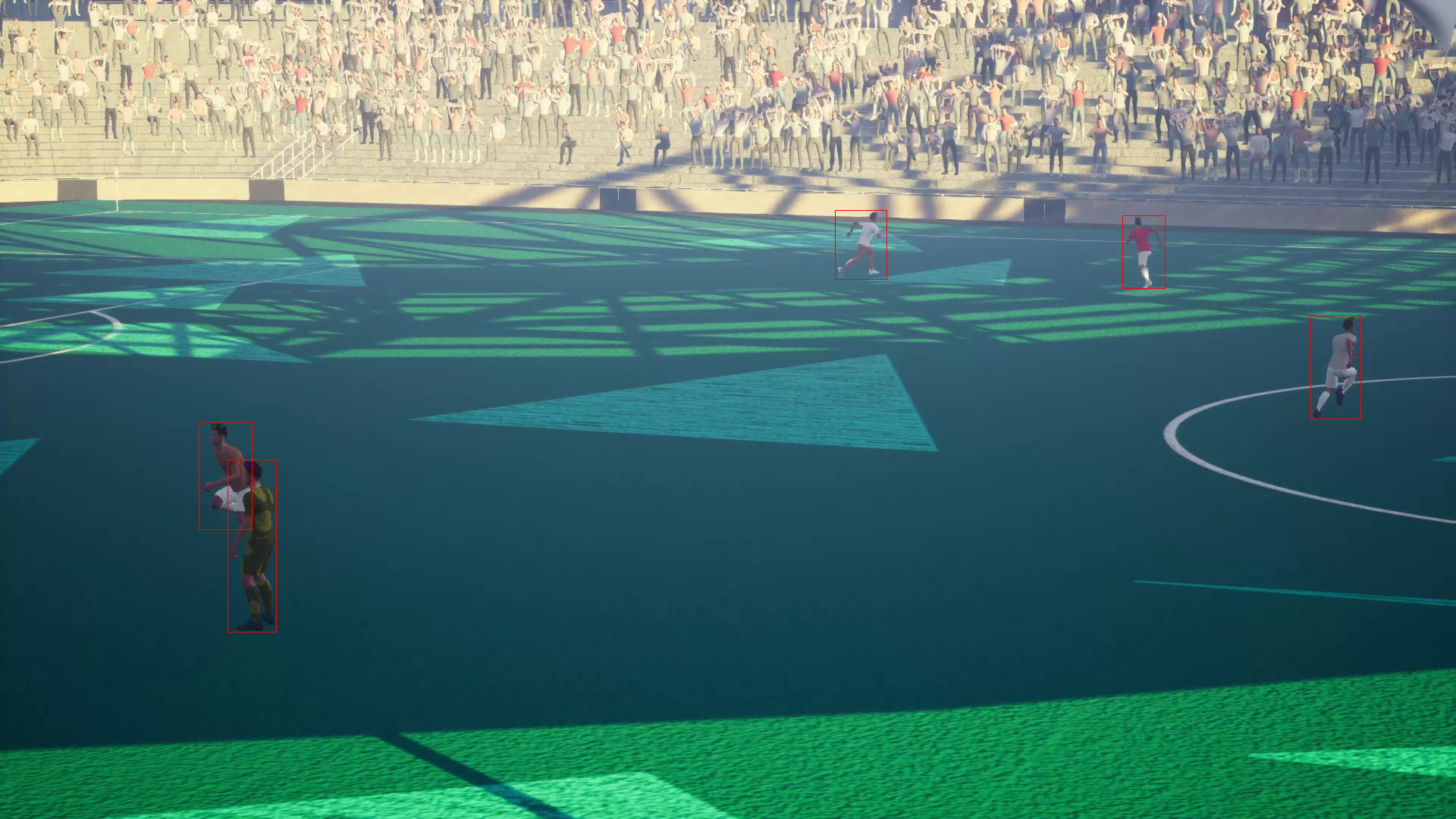}\par\smallskip{\small (c) Lighting}
\end{minipage}
\caption{Examples of SoccerSynth-Detection with motion blur, field-appearance variation, and lighting variation.}
\label{fig:3.2}
\end{figure}

Player motion adds variation in poses and spatial arrangements. We generate camera-motion blur by combining camera orientation perturbations with Unreal Engine Post Process Volume settings. Appendix~\ref{sec:A.1} gives the parameters; Figure~\ref{fig:3.2} illustrates generated variation.

The initial exports contain class 0 for players and class 1 for the ball. For the player-detection experiments, we removed ball annotations after export and retained only class 0. The experimental dataset contains 54,673 images: 36,082 clear images at 1920$\times$1080 and 18,591 motion-blurred images at 800$\times$600. The training and synthetic-validation composition is given in Appendix~\ref{sec:A.6}. The two groups differ in native resolution and aspect ratio as well as motion blur.

\subsection{Experimental Setup and Evaluation Protocol}
\label{sec:3.1.3}

We evaluated the nano and large variants of YOLO version 8 (YOLOv8n and YOLOv8l)~\cite{ultralytics2023} and the large Real-Time Detection Transformer (RT-DETR-L)~\cite{zhao2023detrs} on SoccerNet-Tracking~\cite{cioppa2022soccernet} and the soccer subset of SportsMOT~\cite{cui2023sportsmot}. SoccerSynth-Detection provides 43,734 training images and 10,939 synthetic validation images. Detector initialization uses official weights pretrained on Common Objects in Context (COCO).

\paragraph*{Target-domain data partitions.} We hold out entire matches or source videos from the real training sets for validation. For SoccerNet-Tracking, all clips from match group 4 (SNMOT-060--077) form the validation set, while match groups 6 and 9 form the training set. For SportsMOT, clips c003, c012, and c013 from source video \texttt{v\_gQNyhv8y0QY} form the validation set; the other three source videos remain in the training set. The resulting training/validation/test counts are 29,250/13,500/36,750 for SoccerNet-Tracking and 8,213/2,900/9,058 for SportsMOT. Final testing uses the official SoccerNet-Tracking test and SportsMOT validation partitions. Appendix~\ref{sec:A.2} lists the data composition.

\paragraph*{Training procedures.} Real Only uses the real training partition. Pretraining and Fine-Tuning first trains on SoccerSynth-Detection and then fine-tunes on real training images, using the corresponding same-seed synthetic checkpoint. Mixed Training combines the real training partition with an equal number of SoccerSynth training images. Real Only and Mixed Training start from the corresponding COCO-pretrained weights. Data partitions and image lists are fixed across training seeds 32, 115, and 3407.

\paragraph*{Model selection and evaluation.} We select checkpoints using validation fitness, $0.1\,\mathrm{AP50}+0.9\,\mathrm{AP50{:}95}$, and evaluate the selected \texttt{best.pt} on the test partition. Synthetic pretraining uses the SoccerSynth validation partition; target-domain training uses the real validation partition. Training runs for up to 150 epochs with early-stopping patience 30. We report average precision (AP): AP50 uses an intersection-over-union (IoU) threshold of 0.5, and AP50:95 averages over thresholds 0.50--0.95 in steps of 0.05. Repeated-run results use the mean and sample standard deviation (SD). All groups contain three seeds. Two three-seed groups include a run with a different learning rate and precision setting; these exceptions are marked in the tables. Paired gains use the seeds shared by the compared conditions. Appendix~\ref{sec:A.3} specifies seed membership, and Appendix~\ref{sec:A.4} gives the training and test settings.

\paragraph*{Transfer protocol.} The Transfer Experiment uses the official dataset splits. A detector is trained in a source domain, selected using its source-domain selection set, and evaluated on a different real-image dataset. Real-source training uses 42,750 SoccerNet-Tracking images or 11,113 SportsMOT soccer images, with the source dataset's official evaluation partition used for source-domain checkpoint selection. Synthetic-source training uses the SoccerSynth training and validation partitions. The COCO-only reference is evaluated directly from the official pretrained weights.

For all experiments, retained annotations and predictions are mapped to the single player class. SoccerNet-Tracking labels include the non-ball source annotations, including referees and goalkeepers; SportsMOT uses the people in its original annotations. COCO-only predictions are filtered to person class 0. The synthetic-label preprocessing adds $4/640$ to normalized box width while preserving center and height. Appendix~\ref{sec:A.5} describes these annotation conventions. The input-size parameter is 640, with the detector's standard resizing and postprocessing.

\subsection{Real-Image Augmentation Controls}
\label{sec:3.1.4}

To test whether transformations of real training images can substitute for simulation-generated data, we compare Mixed Training with three real-image augmentation procedures: blur, overlap, and their combination. If the real training partition contains $N$ images, Mixed Training adds $N$ SoccerSynth images, while Real + Blur, Real + Overlap, and Real + Blur + Overlap each add $N$ augmentation entries. The combined condition allocates approximately half of these entries to each augmentation. All source frames and donor players come from the corresponding real training partition.

Blur augmentation applies normalized linear motion kernels of lengths 9, 13, 17, 21, or 25 pixels, with direction sampled from $[0,180)$ degrees; source labels are retained. Overlap augmentation uses Segment Anything Model (SAM) 2~\cite{ravi2025sam2} to extract donor players and paste them into training frames, updating annotations and checking donor identity and placement geometry. The controls match the number of training-list entries. Appendix~\ref{sec:A.6} gives generation settings and image-list composition.

\section{Experiments and Results}
\label{sec:3.2}

Results are aggregated at full precision and rounded to two decimal places. Transfer results follow the source-domain protocol; the remaining comparisons use the fixed target-domain partitions defined in Section~\ref{sec:3.1.3}.

\subsection{Transfer Experiment}
\label{sec:3.2.1}

The relative performance of direct transfer varies by target dataset and detector. Table~\ref{tbl:3.1} compares source-trained checkpoints and the COCO-pretrained evaluation-only baseline on the complete target evaluation sets. On SoccerNet-Tracking, SoccerSynth-Detection pretraining achieved higher AP50:95 than the COCO-only baseline for YOLOv8n and YOLOv8l, but lower AP50:95 for RT-DETR-L. On SportsMOT, it produced lower AP50:95 than the COCO-only baseline for all three detectors.

\par\noindent\begin{minipage}{\linewidth}
\centering
\footnotesize
\captionsetup{singlelinecheck=false}
\setlength{\tabcolsep}{3pt}
\renewcommand{\arraystretch}{1.22}

\captionof{table}[Results of the Transfer Experiment for the three detectors.]{Results of the Transfer Experiment for the three detectors. Each cell reports AP50 / AP50:95 (\%). Bold indicates the highest AP50:95 within each target-detector group, including ties. COCO rows report results obtained with COCO-pretrained weights without additional soccer-domain training.}\label{tbl:3.1}
\begin{tabular}{@{}>{\raggedright\arraybackslash\leavevmode}p{\dimexpr 0.210000\linewidth-1.680000\tabcolsep\relax}>{\raggedright\arraybackslash\leavevmode}p{\dimexpr 0.280000\linewidth-2.240000\tabcolsep\relax}>{\centering\arraybackslash\leavevmode}p{\dimexpr 0.170000\linewidth-1.360000\tabcolsep\relax}>{\centering\arraybackslash\leavevmode}p{\dimexpr 0.170000\linewidth-1.360000\tabcolsep\relax}>{\centering\arraybackslash\leavevmode}p{\dimexpr 0.170000\linewidth-1.360000\tabcolsep\relax}@{}}
\toprule
\textbf{Target dataset} & \textbf{Source training data} & \textbf{YOLOv8n} & \textbf{YOLOv8l} & \textbf{RT-DETR-L} \\[3pt]
\midrule
SoccerNet-Tracking & COCO & 83.82 / 42.30 & 89.99 / 52.30 & \textbf{89.87 / 50.34} \\[3pt]
SoccerNet-Tracking & SoccerSynth-Detection & \textbf{89.68 / 50.11} & \textbf{92.62 / 52.83} & 86.42 / 44.17 \\[3pt]
SoccerNet-Tracking & SportsMOT & 81.10 / 43.33 & 87.17 / 51.08 & 86.83 / 49.27 \\[3pt]
SportsMOT & COCO & 85.32 / 58.04 & \textbf{90.62 / 73.39} & \textbf{89.84 / 68.37} \\[3pt]
SportsMOT & SoccerSynth-Detection & 87.23 / 55.78 & 88.77 / 59.84 & 83.54 / 44.02 \\[3pt]
SportsMOT & SoccerNet-Tracking & \textbf{90.25 / 62.44} & 90.64 / 65.63 & 90.73 / 63.83 \\[3pt]
\bottomrule
\end{tabular}
\end{minipage}\par

\begin{figure}[!htbp]
\centering
\begin{minipage}[b]{0.32\linewidth}\centering
\includegraphics[width=\linewidth,height=0.25\textheight,keepaspectratio]{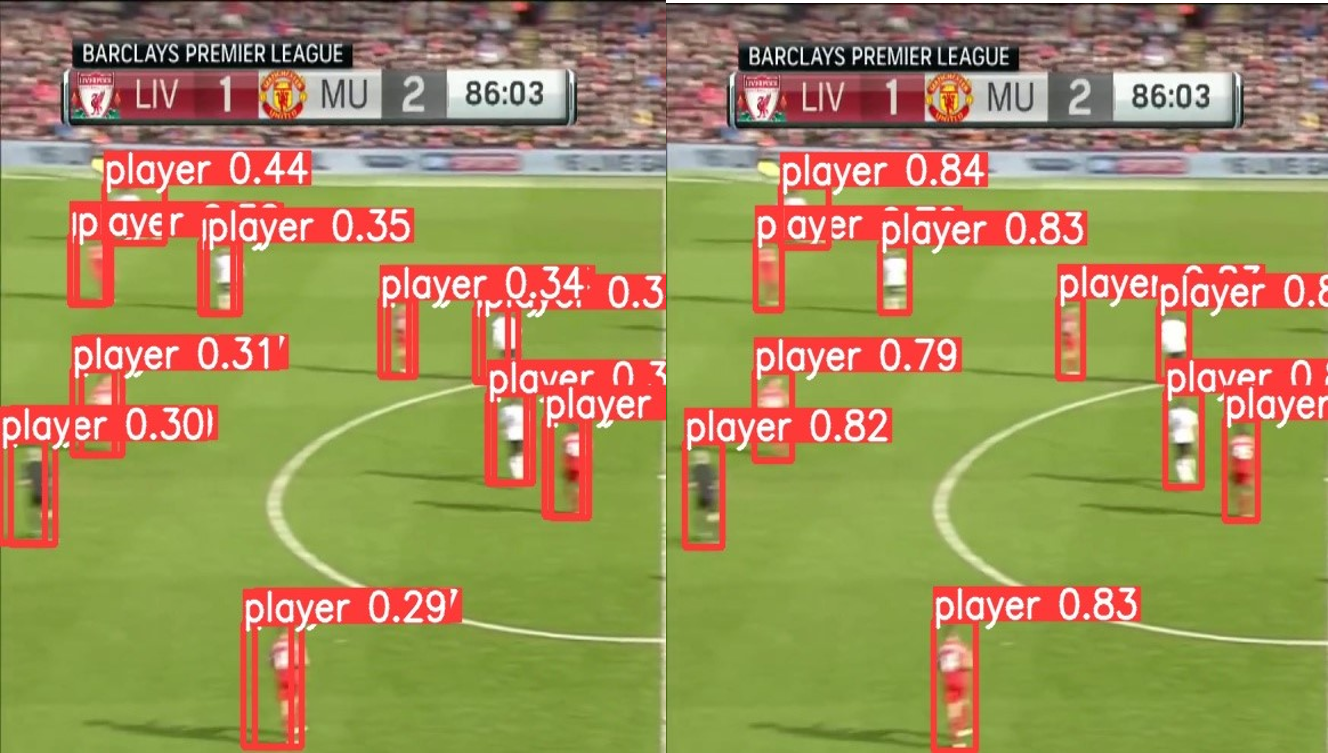}\par\smallskip{\small (a)}
\end{minipage}
\hfill
\begin{minipage}[b]{0.32\linewidth}\centering
\includegraphics[width=\linewidth,height=0.25\textheight,keepaspectratio]{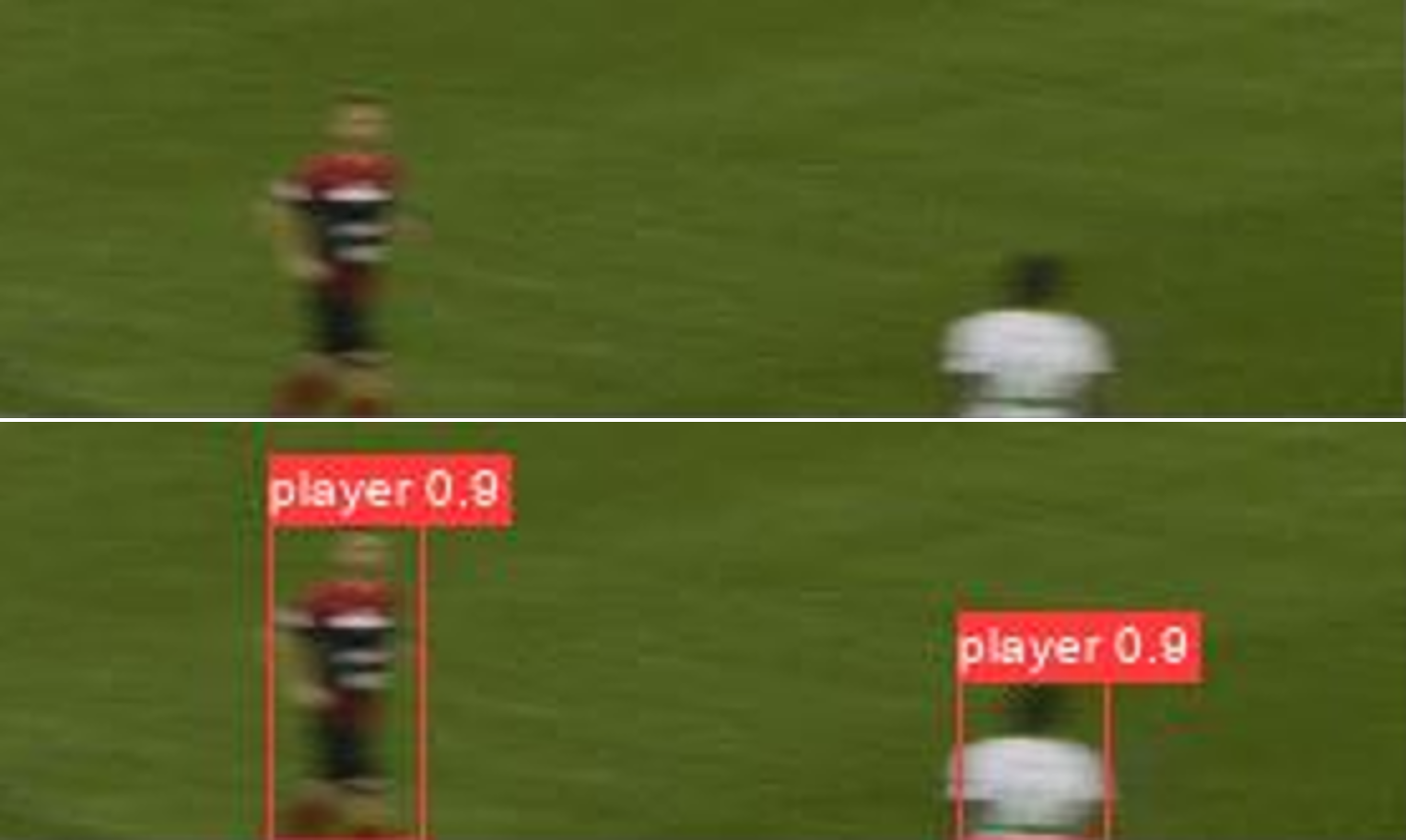}\par\smallskip{\small (b)}
\end{minipage}
\hfill
\begin{minipage}[b]{0.32\linewidth}\centering
\includegraphics[width=\linewidth,height=0.25\textheight,keepaspectratio]{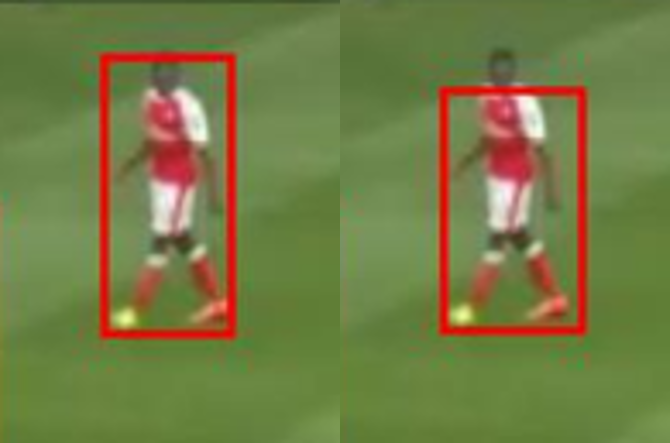}\par\smallskip{\small (c)}
\end{minipage}
\caption[YOLOv8n qualitative examples from the Transfer Experiment.]{YOLOv8n qualitative examples from the Transfer Experiment. From left to right: (a) results obtained with SoccerNet-Tracking (left) and SoccerSynth-Detection (right) on a motion-blurred frame; (b) results obtained with SportsMOT (top) and SoccerSynth-Detection (bottom) on a motion-blurred frame; and (c) results obtained with SoccerNet-Tracking (left) and SoccerSynth-Detection (right) in a performance-degradation case.}
\label{fig:3.4}
\end{figure}

The qualitative examples in Figure~\ref{fig:3.4} show detections under blur and crowded-player conditions, but also missed head regions and incomplete boxes. These examples suggest limitations in appearance and viewpoint coverage, although they do not isolate the causes of the errors.

\subsection{Training Procedure Comparison}
\label{sec:3.2.2}

Mixed Training improves mean AP50:95 over Real Only in all six detector--dataset combinations when the same seeds are compared (Table~\ref{tbl:3.2}). The paired mean gains range from 1.43 to 3.38 percentage points. Pretraining and Fine-Tuning also improves mean AP50:95 in all six combinations, with paired gains of 0.22--3.37 percentage points. Fine-tuning has a higher mean than Mixed Training for YOLOv8n and YOLOv8l on SportsMOT, illustrating the dependence on training procedure and target dataset. Paired gains are calculated over the three matched seeds before rounding and may therefore differ slightly from differences between the displayed rounded means.

\par\medskip\noindent\begin{minipage}{\linewidth}

\captionsetup{labelfont={color=black},textfont={color=black}}
\centering\footnotesize
\captionsetup{singlelinecheck=false,width=\linewidth}
\setlength{\tabcolsep}{6pt}
\renewcommand{\arraystretch}{1.08}
\captionof{table}{Detection performance by target-domain training procedure. AP50 is reported as the mean; AP50:95 is reported as the mean $\pm$ sample SD (\%). Results use three seeds; $\ddagger$ marks a group containing one learning-rate/AMP exception (Appendices~\ref{sec:A.3}--\ref{sec:A.4}).}\label{tbl:3.2}
\begin{tabular}{@{}llccc@{}}
\toprule
\textbf{Detector} & \textbf{Metric} & \shortstack{Real\\Only} & \shortstack{Pretraining and\\Fine-Tuning} & \shortstack{Mixed\\Training} \\[3pt]
\midrule
\multicolumn{5}{@{}l}{\textbf{SoccerNet-Tracking}} \\[3pt]
\raisebox{-5.5pt}[0pt][0pt]{YOLOv8n} & AP50 & 94.82 & 96.09 & 96.73 \\
 & AP50:95 & $58.54\,\pm\,0.59$ & $61.03\,\pm\,0.16$ & $61.92\,\pm\,0.09$ \\[6pt]
\raisebox{-5.5pt}[0pt][0pt]{YOLOv8l} & AP50 & 97.16 & 97.21 & 97.64 \\
 & AP50:95 & $64.56\,\pm\,0.17$ & $64.78\,\pm\,0.32$ & $65.99\,\pm\,0.11$ \\[6pt]
\raisebox{-5.5pt}[0pt][0pt]{RT-DETR-L} & AP50 & 95.66 & 95.88 & 96.59 \\
 & AP50:95 & $60.10\,\pm\,0.60$ & $61.11\,\pm\,0.33$ & $61.67\,\pm\,0.46^{\ddagger}$ \\[6pt]
\midrule
\multicolumn{5}{@{}l}{\textbf{SportsMOT soccer}} \\[3pt]
\raisebox{-5.5pt}[0pt][0pt]{YOLOv8n} & AP50 & 90.93 & 91.84 & 91.72 \\
 & AP50:95 & $72.72\,\pm\,1.47$ & $76.09\,\pm\,0.61$ & $75.87\,\pm\,0.25$ \\[6pt]
\raisebox{-5.5pt}[0pt][0pt]{YOLOv8l} & AP50 & 92.51 & 94.64 & 93.60 \\
 & AP50:95 & $79.14\,\pm\,1.43$ & $81.68\,\pm\,0.53$ & $81.49\,\pm\,1.12$ \\[6pt]
\raisebox{-5.5pt}[0pt][0pt]{RT-DETR-L} & AP50 & 92.91 & 92.80 & 93.67 \\
 & AP50:95 & $76.82\,\pm\,0.85$ & $78.60\,\pm\,1.17$ & $79.75\,\pm\,0.15$ \\[6pt]
\bottomrule
\end{tabular}
\end{minipage}\par

\subsection{Real-Image Augmentation Comparison}
\label{sec:3.2.3}

Mixed Training has higher mean AP50:95 than each real-image augmentation control in all six detector--dataset combinations (Table~\ref{tbl:3.3}). All conditions use three seeds, with the SoccerNet-Tracking RT-DETR-L configuration exceptions marked in the table. Thus, under the evaluated conditions, none of the three augmentation procedures matches the detection performance obtained by adding synthetic images. This comparison tests their ability to substitute for synthetic data; it does not isolate the visual factors responsible for the gains.

\par\medskip\noindent\begin{minipage}{\linewidth}

\captionsetup{labelfont={color=black},textfont={color=black}}
\centering\footnotesize
\captionsetup{singlelinecheck=false,width=\linewidth}
\setlength{\tabcolsep}{6pt}
\renewcommand{\arraystretch}{1.08}
\captionof{table}{Detection performance with synthetic data and real-image augmentation. AP50 is reported as the mean; AP50:95 is reported as the mean $\pm$ sample SD (\%). Results use three seeds; $\ddagger$ marks a group containing one learning-rate/AMP exception (Appendices~\ref{sec:A.3}--\ref{sec:A.4}).}\label{tbl:3.3}
\begin{tabular}{@{}llcccc@{}}
\toprule
\textbf{Detector} & \textbf{Metric} & \shortstack{Mixed\\Training} & \shortstack{Real +\\Blur} & \shortstack{Real +\\Overlap} & \shortstack{Real + Blur\\+ Overlap} \\[3pt]
\midrule
\multicolumn{6}{@{}l}{\textbf{SoccerNet-Tracking}} \\[3pt]
\raisebox{-5.5pt}[0pt][0pt]{YOLOv8n} & AP50 & 96.73 & 94.99 & 95.51 & 95.22 \\
 & AP50:95 & $61.92\,\pm\,0.09$ & $59.03\,\pm\,0.14$ & $59.22\,\pm\,0.08$ & $59.76\,\pm\,0.28$ \\[6pt]
\raisebox{-5.5pt}[0pt][0pt]{YOLOv8l} & AP50 & 97.64 & 96.80 & 97.24 & 97.05 \\
 & AP50:95 & $65.99\,\pm\,0.11$ & $63.47\,\pm\,0.73$ & $64.53\,\pm\,0.11$ & $64.33\,\pm\,0.24$ \\[6pt]
\raisebox{-5.5pt}[0pt][0pt]{RT-DETR-L} & AP50 & 96.59 & 95.38 & 95.80 & 95.52 \\
 & AP50:95 & $61.67\,\pm\,0.46^{\ddagger}$ & $59.88\,\pm\,0.60$ & $60.80\,\pm\,0.94$ & $60.16\,\pm\,0.69^{\ddagger}$ \\[6pt]
\midrule
\multicolumn{6}{@{}l}{\textbf{SportsMOT soccer}} \\[3pt]
\raisebox{-5.5pt}[0pt][0pt]{YOLOv8n} & AP50 & 91.72 & 92.45 & 92.02 & 91.91 \\
 & AP50:95 & $75.87\,\pm\,0.25$ & $75.08\,\pm\,0.28$ & $75.08\,\pm\,0.20$ & $74.60\,\pm\,0.15$ \\[6pt]
\raisebox{-5.5pt}[0pt][0pt]{YOLOv8l} & AP50 & 93.60 & 93.49 & 92.91 & 93.40 \\
 & AP50:95 & $81.49\,\pm\,1.12$ & $80.36\,\pm\,0.62$ & $79.59\,\pm\,0.59$ & $79.52\,\pm\,0.65$ \\[6pt]
\raisebox{-5.5pt}[0pt][0pt]{RT-DETR-L} & AP50 & 93.67 & 92.31 & 92.91 & 93.21 \\
 & AP50:95 & $79.75\,\pm\,0.15$ & $75.35\,\pm\,1.84$ & $77.07\,\pm\,0.69$ & $77.26\,\pm\,0.87$ \\[6pt]
\bottomrule
\end{tabular}
\end{minipage}\par

\subsection{Analysis of Synthetic Data Proportions}
\label{sec:3.2.4}

We fix the real training partition and compare real-to-synthetic ratios of 1:0, 1:0.5, 1:1, and approximately 1:1.5. The synthetic subsets are nested and drawn from the existing 43,734-image SoccerSynth training pool. SoccerNet-Tracking uses 14,625, 29,250, or 43,734 synthetic images; SportsMOT uses 4,107, 8,213, or 12,320. The highest SoccerNet-Tracking ratio is approximately 1:1.495.

Table~\ref{tbl:3.4} shows that the preferred proportion varies by detector and dataset. YOLOv8n benefits from the highest tested proportion on both datasets. For SportsMOT YOLOv8l, the mean at 1:1 exceeds that at approximately 1:1.5, while RT-DETR-L has its highest reported mean at 1:0.5. The proportion comparisons therefore describe a range of useful mixtures rather than a single optimum shared by all models.

\par\medskip\noindent\begin{minipage}{\linewidth}

\captionsetup{labelfont={color=black},textfont={color=black}}
\centering\footnotesize
\captionsetup{singlelinecheck=false,width=\linewidth}
\setlength{\tabcolsep}{6pt}
\renewcommand{\arraystretch}{1.08}
\captionof{table}{Detection performance at different real-to-synthetic ratios. AP50 is reported as the mean; AP50:95 is reported as the mean $\pm$ sample SD (\%). Results use three seeds; $\ddagger$ marks a group containing one learning-rate/AMP exception (Appendices~\ref{sec:A.3}--\ref{sec:A.4}).}\label{tbl:3.4}
\begin{tabular}{@{}llcccc@{}}
\toprule
\textbf{Detector} & \textbf{Metric} & 1:0 & 1:0.5 & 1:1 & $\approx$1:1.5 \\[3pt]
\midrule
\multicolumn{6}{@{}l}{\textbf{SoccerNet-Tracking}} \\[3pt]
\raisebox{-5.5pt}[0pt][0pt]{YOLOv8n} & AP50 & 94.82 & 96.60 & 96.73 & 96.86 \\
 & AP50:95 & $58.54\,\pm\,0.59$ & $61.57\,\pm\,0.33$ & $61.92\,\pm\,0.09$ & $62.34\,\pm\,0.13$ \\[6pt]
\raisebox{-5.5pt}[0pt][0pt]{YOLOv8l} & AP50 & 97.16 & 97.49 & 97.64 & 97.56 \\
 & AP50:95 & $64.56\,\pm\,0.17$ & $65.37\,\pm\,0.13$ & $65.99\,\pm\,0.11$ & $66.05\,\pm\,0.04$ \\[6pt]
\raisebox{-5.5pt}[0pt][0pt]{RT-DETR-L} & AP50 & 95.66 & 96.43 & 96.59 & 96.66 \\
 & AP50:95 & $60.10\,\pm\,0.60$ & $61.74\,\pm\,0.07$ & $61.67\,\pm\,0.46^{\ddagger}$ & $61.93\,\pm\,0.79$ \\[6pt]
\midrule
\multicolumn{6}{@{}l}{\textbf{SportsMOT soccer}} \\[3pt]
\raisebox{-5.5pt}[0pt][0pt]{YOLOv8n} & AP50 & 90.93 & 91.45 & 91.72 & 92.16 \\
 & AP50:95 & $72.72\,\pm\,1.47$ & $75.43\,\pm\,0.49$ & $75.87\,\pm\,0.25$ & $76.73\,\pm\,0.14$ \\[6pt]
\raisebox{-5.5pt}[0pt][0pt]{YOLOv8l} & AP50 & 92.51 & 93.50 & 93.60 & 93.54 \\
 & AP50:95 & $79.14\,\pm\,1.43$ & $81.17\,\pm\,0.68$ & $81.49\,\pm\,1.12$ & $80.92\,\pm\,1.93$ \\[6pt]
\raisebox{-5.5pt}[0pt][0pt]{RT-DETR-L} & AP50 & 92.91 & 94.44 & 93.67 & 93.30 \\
 & AP50:95 & $76.82\,\pm\,0.85$ & $79.80\,\pm\,0.47$ & $79.75\,\pm\,0.15$ & $79.30\,\pm\,0.33$ \\[6pt]
\bottomrule
\end{tabular}
\end{minipage}\par

\section{Discussion}
\label{sec:3.3}

\subsection{Synthetic Data as Supplementary Training Data}
\label{sec:3.3.1}

The Transfer Experiment shows that synthetic-only training has different outcomes across real target datasets. Combining synthetic and real training data, either through mixed training or pretraining followed by fine-tuning, improves detection performance in the target-domain comparisons. The augmentation comparison tests whether the specified real-image transformations can replace synthetic training samples, while the proportion study examines how much synthetic data to add.

SoccerSynth varies player appearance, pose, scale, spatial configuration, camera conditions, and background together~\cite{prakash2019structured,fabbri2021motsynth}. These variations offer a possible explanation for how generated samples complement real training images. The experiments establish the benefit of the complete training procedures, but do not determine the contribution of each visual factor.

\subsection{Evaluation Scope}
\label{sec:3.3.2}

The evaluation covers single-image player detection on two broadcast-image datasets. SoccerNet-Tracking partitions are separated by parent match and SportsMOT partitions by source video. Sample SD describes variation across the reported runs, including the marked configuration exceptions. The procedures differ in image counts, pretraining stages, and early-stopping points, which also affect training updates.

SoccerSynth-Detection provides center-camera images rather than continuous tracking sequences. Its randomization covers a limited range of broadcast appearances; weather, camera color responses, compression artifacts, and lens effects could be extended. The source datasets also differ in annotated roles and bounding-box conventions (Appendix~\ref{sec:A.5}).

\section{Conclusion}
\label{sec:3.4}

SoccerSynth-Detection provides annotated soccer images for detector training. The Transfer Experiment examines source-to-target application using the official dataset splits. In the target-domain experiments, both mixed training and synthetic pretraining followed by fine-tuning improve mean AP50:95 over matched real-only seeds in all six detector--dataset combinations. The specified real-image augmentations do not match the performance of mixed training, while the proportion comparisons show that the preferred amount of synthetic data varies by detector and dataset.

\section*{Funding}
This work was financially supported by JST SPRING, Grant Number JPMJSP2125. The author H. Q. would like to take this opportunity to thank the ``THERS Make New Standards Program for the Next Generation Researchers''.

\vspace{1em}
\noindent \textbf{Data availability statement}
The SoccerSynth-Detection dataset generated in this study is archived on Zenodo at \url{https://doi.org/10.5281/zenodo.20839314}. The third-party datasets analyzed in this study are publicly available from their original sources: SoccerNet-Tracking (dataset paper DOI: \url{https://doi.org/10.1109/CVPRW56347.2022.00393}) and SportsMoT (dataset paper DOI: \url{https://doi.org/10.1109/ICCV51070.2023.00910}). SoccerNet-Tracking was accessed through the official SoccerNet data distribution, and SportsMoT was accessed from the official dataset repository/page. Users should follow the licenses, permissions, and access conditions specified by the original dataset providers.

\vspace{1em}
\noindent \textbf{Code availability statement}
Version 1.0.0 of the SoccerSynth-Detection experiment artifacts is archived on Zenodo at \url{https://doi.org/10.5281/zenodo.21021022}. The corresponding GitHub repository is available at \url{https://github.com/open-starlab/SoccerSynth-Detection}.

\section*{Declarations}
\vspace{1em}
\noindent \textbf{Conflict of Interest}
All authors declare that they have no conflict of interest.

\noindent \textbf{Author Contribution}
H.Q. implemented the system and H.Q. and K.F. wrote the main manuscript text. All authors discussed this research and reviewed the manuscript.

\clearpage\appendix
\section{Randomization Factors in SoccerSynth-Detection}
\label{sec:A.1}

\begin{table}[!htbp]
\centering
\renewcommand{\baselinestretch}{1.15}
\footnotesize
\captionsetup{singlelinecheck=false}
\setlength{\tabcolsep}{3pt}
\renewcommand{\arraystretch}{1.05}

\caption[SoccerSynth randomization factors.]{Randomization Factors in SoccerSynth-Detection.}\label{tbl:A.1}
\begin{tabular}{@{}>{\raggedright\arraybackslash\leavevmode}p{\dimexpr 0.170000\linewidth-0.680000\tabcolsep\relax}>{\raggedright\arraybackslash\leavevmode}p{\dimexpr 0.380000\linewidth-1.520000\tabcolsep\relax}>{\raggedright\arraybackslash\leavevmode}p{\dimexpr 0.450000\linewidth-1.800000\tabcolsep\relax}@{}}
\toprule
\textbf{Factor} & \textbf{Implementation/settings} & \textbf{Scope/limitations} \\[3pt]
\midrule
Motion blur & Camera-orientation perturbation plus Post Process Volume motion blur: amount 1.0, maximum 9.12, target frame rate 0 frames per second (FPS), per-object size 7.04. & Clear images: 1920$\times$1080; motion-blurred images: 800$\times$600. \\[3pt]
Field appearance & Field material sampled from array index 0--90 and applied to field mesh slot 0. & Circle, diamond, triangle, stripe, clear/plain, and realistic field textures. \\[3pt]
Lighting & Directional sunlight: pitch 180--360 degrees, yaw 0--360 degrees, intensity 0--1 lux, B channel 0--1 (R=G=1). & Intensity values are assigned directly to the light, without scaling or normalization. Daylight-style variation only; no night-time stadium floodlights or weather effects. \\[3pt]
Player appearance & Clothing/uniform, shoe, and hair materials sampled from indices 1--55 and assigned to slots 1, 2, and 5. & Solid, dot, block, stripe, and two-tone styles; no logos, numbers, or accessories. \\[3pt]
Camera settings & Standard UE camera aimed at a random target point in world coordinates (cm): X=-5260--5250, Y=-2660--2660, Z=0; default horizontal field of view (FOV) 90 degrees, with an offset of $-20$ to $-10$ degrees (resulting FOV: 70--80 degrees). & Broadcast-style center-camera views, with no additional coordinate scaling. \\[3pt]
Player actions and scene recreation & Behavior Tree and AI Controller components simulate running, passing, shooting, receiving, and repositioning in autonomous match-like sequences. & Dynamic frame-level scenes; generated frames are not identity-continuous tracking sequences. \\[3pt]
\bottomrule
\end{tabular}
\end{table}

\clearpage
\section{Data Partitions and Training-Set Composition}
\label{sec:A.2}

Table~\ref{tbl:A.2} gives the target-domain data partitions. SoccerNet-Tracking uses match groups 6 and 9 for training, group 4 for validation, and groups 7, 8, and 11 for testing. The validation group contains clips SNMOT-060--077.

SportsMOT validation comprises clips c003, c012, and c013 from source video \texttt{v\_gQNyhv8y0QY}. Training uses source videos \texttt{v\_1yHWGw8DH4A}, \texttt{v\_HdiyOtliFiw}, and \texttt{v\_iIxMOsCGH58}. Testing uses the five videos in the official validation partition: \texttt{v\_2QhNRucNC7E}, \texttt{v\_G-vNjfx1GGc}, \texttt{v\_ITo3sCnpw\_k}, \texttt{v\_dw7LOz17Omg}, and \texttt{v\_i2\_L4qquVg0}. Group identifiers and exact image hashes are disjoint across partitions. SportsMOT grouping establishes separation at the source-video level.

\par\medskip\noindent\begin{minipage}{\linewidth}

\captionsetup{labelfont={color=black},textfont={color=black}}
\centering\footnotesize
\captionsetup{singlelinecheck=false,width=\linewidth}
\setlength{\tabcolsep}{5pt}
\renewcommand{\arraystretch}{1.15}
\captionof{table}{Real-image partitions for the target-domain experiments.}\label{tbl:A.2}
\begin{tabular}{@{}lrrr@{}}
\toprule
\textbf{Dataset} & \textbf{Training} & \textbf{Validation} & \textbf{Test} \\[3pt]
\midrule
SoccerNet-Tracking & 29,250 & 13,500 & 36,750 \\[3pt]
SportsMOT soccer & 8,213 & 2,900 & 9,058 \\[3pt]
\bottomrule
\end{tabular}
\end{minipage}\par

The real partitions and training lists are fixed across seeds. The 1:0.5, 1:1, and approximately 1:1.5 mixtures add nested SoccerSynth subsets of 14,625, 29,250, and 43,734 images for SoccerNet-Tracking, and 4,107, 8,213, and 12,320 for SportsMOT. All synthetic samples come from the SoccerSynth training partition.

\section{Checkpoint Selection and Statistical Aggregation}
\label{sec:A.3}

Target-domain checkpoints are selected by fitness on the real validation partition, $0.1\,\mathrm{AP50}+0.9\,\mathrm{AP50{:}95}$, and evaluated on the test partition. Synthetic pretraining uses the synthetic validation partition for selection. Transfer checkpoints follow the source-domain protocol in Section~\ref{sec:3.1.3}.

Statistics use full-precision test outputs where available; the SoccerNet-Tracking RT-DETR-L 1:1.5 seed-115 metrics are available to two decimal places in percentage units. Mean and sample SD are computed across the reported runs for each condition; sample variance uses denominator $n-1$. Table~\ref{tbl:A.3} lists the groups with configuration exceptions. Paired gains use seeds common to the compared conditions. The three-seed aggregates include the two completed learning-rate/AMP exception runs described in Appendix~\ref{sec:A.4}; their SD therefore reflects variation across these runs rather than repetitions of an identical configuration.

\par\medskip\noindent\begin{minipage}{\linewidth}

\captionsetup{labelfont={color=black},textfont={color=black}}
\centering\footnotesize
\captionsetup{singlelinecheck=false,width=\linewidth}
\setlength{\tabcolsep}{5pt}
\renewcommand{\arraystretch}{1.3}
\captionof{table}{Groups with configuration exceptions. All groups use seeds 32, 115, and 3407. $\ddagger$ marks a group containing one learning-rate/AMP exception; other groups use the standard configuration.}\label{tbl:A.3}
\begin{tabular*}{\linewidth}{@{\extracolsep{\fill}}lll@{}}
\toprule
\textbf{Dataset / detector} & \textbf{Condition} & \textbf{Seeds} \\[3pt]
\midrule
\shortstack[l]{SoccerNet-Tracking\\RT-DETR-L} & Mixed Training $\ddagger$ & 32, 115, 3407 \\[3pt]
\shortstack[l]{SoccerNet-Tracking\\RT-DETR-L} & Real + Blur + Overlap $\ddagger$ & 32, 115, 3407 \\[3pt]
\bottomrule
\end{tabular*}
\end{minipage}\par

\section{Training and Test Settings}
\label{sec:A.4}

Table~\ref{tbl:A.4} summarizes the standard real-domain configurations. Real Only and Mixed Training start from official COCO-pretrained weights; Pretraining and Fine-Tuning starts from the corresponding same-seed SoccerSynth checkpoint. Synthetic pretraining uses 43,734 training and 10,939 validation images.

\par\medskip\noindent\begin{minipage}{\linewidth}

\captionsetup{labelfont={color=black},textfont={color=black}}
\centering\footnotesize
\captionsetup{singlelinecheck=false,width=\linewidth}
\setlength{\tabcolsep}{5pt}
\renewcommand{\arraystretch}{1.15}
\captionof{table}{Standard target-domain training configurations. Batch sizes are fixed across target-domain training conditions within each dataset. All standard configurations use AMP, at most 150 epochs, and early-stopping patience 30.}\label{tbl:A.4}
\begin{tabular*}{\linewidth}{@{\extracolsep{\fill}}lcclc@{}}
\toprule
\textbf{Detector} & \shortstack{\textbf{Batch size}\\\textbf{SoccerNet}} & \shortstack{\textbf{Batch size}\\\textbf{SportsMOT}} & \textbf{Optimizer} & \textbf{Learning rate} \\[3pt]
\midrule
YOLOv8n & 64 & 64 & SGD & 0.01 \\[3pt]
YOLOv8l & 16 & 32 & SGD & 0.01 \\[3pt]
RT-DETR-L & 32 & 32 & AdamW & $10^{-4}$ \\[3pt]
\bottomrule
\end{tabular*}
\end{minipage}\par\medskip

Two SoccerNet-Tracking RT-DETR-L runs use learning rate $5\times10^{-5}$ with automatic mixed precision (AMP) disabled: Mixed Training seed 3407 achieves AP50:95 61.25, and Real + Blur + Overlap seed 115 achieves 59.86. Both are included in their respective three-seed aggregates, marked $\ddagger$ in the result tables. The other seeds use learning rate $10^{-4}$ with AMP enabled.

The real-domain training records use Ultralytics 8.1.42 and PyTorch 2.2.2+cu121. Final evaluation uses image size 640, batch size 8, confidence threshold 0.001, IoU threshold 0.7, at most 300 detections, and 32-bit floating-point precision (FP32). YOLOv8 uses aspect-ratio-preserving letterboxing and native non-maximum suppression; RT-DETR uses square resizing and query postprocessing.

\section{Annotation Targets and Bounding-Box Conventions}
\label{sec:A.5}

The synthetic player annotation targets are the 22 on-field players placed in the UE scene. The annotation API supplies full-body bounding boxes, so occlusion does not shrink a labeled player's box to its visible portion. The number of exported boxes can vary by image.

Full-body extent under occlusion, image-boundary clipping, and off-screen-object filtering are separate annotation choices. The plugin documentation describes optional padding and a minimum-visible-pixel threshold~\cite{plasma}, but the filtering thresholds and boundary settings are not fully documented for all captures. Before detector-specific loading and geometric processing, 20,190 training boxes and 4,925 synthetic-validation boxes extend beyond the image boundary.

Class mapping is applied after source annotation. The synthetic detection release discards ball class 1 and retains player class 0. SoccerNet-derived labels retain non-ball source annotations, including goalkeeper, referee and other source categories. For SportsMOT soccer, we retain the people annotated in the source dataset without applying an additional role filter. The COCO-only baseline filters predictions to person class 0. Sharing one evaluation class does not establish that all source datasets annotate the same set of people.

The synthetic-label preprocessing code adds a fixed normalized increment of $4/640$ to box width before detector loading and resizing. It transforms a normalized YOLO box $(c_x,c_y,w,h)$ into $(c_x,c_y,w+4/640,h)$, leaving the center and height unchanged. At original image width $W$, the total width increment is $W/160$ pixels: 12 pixels at width 1920, or 5 pixels at width 800. Only at the reference width 640 does it correspond to two pixels on each side. The rationale for this width adjustment is not fully documented.

\section{Dataset Composition and Augmentation Procedures}
\label{sec:A.6}

Table~\ref{tbl:A.5} gives the composition of SoccerSynth-Detection. Training and synthetic-validation image lists are disjoint. The clear and blurred groups also differ in native resolution and aspect ratio.

\begingroup
\footnotesize

\begin{table}[!htbp]
\centering\footnotesize
\setlength{\tabcolsep}{6pt}
\renewcommand{\arraystretch}{1.22}
\begin{lrbox}{\ResultTableBox}
\begin{tabular}{@{}lrrr@{}}
\toprule
Split & Clear & Blurred & Total images \\[3pt]
\midrule
Training & 28,762 & 14,972 & 43,734 \\[3pt]
Synthetic validation & 7,320 & 3,619 & 10,939 \\[3pt]
Total & 36,082 & 18,591 & 54,673 \\[3pt]
\bottomrule
\end{tabular}
\end{lrbox}
\begin{minipage}{\linewidth}
\captionsetup{singlelinecheck=false,width=\linewidth}
\caption[Composition of the synthetic dataset.]{Composition of the 54,673-image experimental dataset. Clear and blurred refer to the recorded source groups, with native sizes 1920$\times$1080 and 800$\times$600, respectively.}\label{tbl:A.5}
\noindent\makebox[\linewidth][c]{\usebox{\ResultTableBox}}
\end{minipage}
\end{table}
\endgroup

Table~\ref{tbl:A.6} specifies the real-image augmentation counts. Generation uses only frames and donor players from the corresponding real training partition. Motion blur uses normalized rotated linear kernels of lengths 9, 13, 17, 21, or 25 pixels, with directions sampled from $[0,180)$ degrees and JPEG quality 95. The generation seeds are 32 for SoccerNet-Tracking and 33 for SportsMOT; source detection boxes are retained.

\par\medskip\noindent\begin{minipage}{\linewidth}

\captionsetup{labelfont={color=black},textfont={color=black}}
\centering\footnotesize
\captionsetup{singlelinecheck=false,width=\linewidth}
\setlength{\tabcolsep}{5pt}
\renewcommand{\arraystretch}{1.15}
\captionof{table}{Additional training entries in the real-image augmentation controls. Every condition also includes the complete real training partition.}\label{tbl:A.6}
\begin{tabular}{@{}lrrrr@{}}
\toprule
\textbf{Dataset} & \shortstack{\textbf{Real}\\\textbf{images}} & \shortstack{\textbf{Blur-only}\\\textbf{entries}} & \shortstack{\textbf{Overlap-only}\\\textbf{entries}} & \shortstack{\textbf{Combined}\\\textbf{blur + overlap}} \\[3pt]
\midrule
SoccerNet-Tracking & 29,250 & 29,250 & 29,250 & 14,625 + 14,625 \\[3pt]
SportsMOT soccer & 8,213 & 8,213 & 8,213 & 4,106 + 4,107 \\[3pt]
\bottomrule
\end{tabular}
\end{minipage}\par

Overlap generation pastes segmented donor players into training frames and adds their detection boxes while retaining destination annotations. Identity checks exclude donor track IDs already present in a destination frame, and geometric checks assess the proposed placement. Alternative training backgrounds supply 1,167 SoccerNet-Tracking slots and 101 SportsMOT slots. The Real + Blur lists contain 201 and 3 repeated image paths, respectively; the combined lists contain 48 and 2. Training-list counts therefore describe sample entries rather than distinct images or backgrounds.

\end{document}